\pdfoutput=1

\documentclass[11pt]{article}

\usepackage[preprint]{acl}

\usepackage{times}
\usepackage{latexsym}

\usepackage[T1]{fontenc}

\usepackage[utf8]{inputenc}

\usepackage{microtype}

\usepackage{inconsolata}

\usepackage{graphicx}
\usepackage{booktabs}
\usepackage{adjustbox}
\usepackage{multirow}
\usepackage{makecell}

\usepackage{CJKutf8}

\title{Towards Spoken Mathematical Reasoning: \\
Benchmarking Speech-based Models over Multi-faceted Math Problems}

\author{
Chengwei Wei\textsuperscript{$\diamondsuit$}, 
Bin Wang\textsuperscript{$\diamondsuit$}, 
Jung-jae Kim \textsuperscript{$\diamondsuit$}, 
Nancy F. Chen\textsuperscript{$\diamondsuit,\dag$}
\\
\textsuperscript{$\diamondsuit$}Institute for Infocomm Research (I$^2$R), A*STAR, Singapore\\
\textsuperscript{$\dag$}Centre for Frontier AI Research (CFAR), A*STAR, Singapore\\
\texttt{wei\_chengwei@i2r.a-star.edu.sg} \\
}

\begin{document}
\maketitle
\begin{abstract}
Recent advances in large language models (LLMs) and multimodal LLMs (MLLMs) have led to strong reasoning ability across a wide range of tasks. However, their ability to perform mathematical reasoning from spoken input remains underexplored. Prior studies on speech modality have mostly focused on factual speech understanding or simple audio reasoning tasks, providing limited insight into logical step-by-step reasoning, such as that required for mathematical problem solving. To address this gap, we introduce Spoken Math Question Answering (Spoken-MQA) \footnote{Our evaluation code and datasets will be open-sourced at \url{https://github.com/amao0o0/Spoken-MQA}}, a new benchmark designed to evaluate the mathematical reasoning capabilities of speech-based models, including both cascade models (ASR + LLMs) and end-to-end speech LLMs. Spoken-MQA covers a diverse set of math problems, including pure arithmetic, single-step and multi-step contextual reasoning, and knowledge-oriented reasoning problems, all presented in unambiguous natural spoken language. Through extensive experiments, we find that: (1) while some speech LLMs perform competitively on contextual reasoning tasks involving basic arithmetic, they still struggle with direct arithmetic problems; (2) current LLMs exhibit a strong bias toward symbolic mathematical expressions written in LaTex and have difficulty interpreting verbalized mathematical expressions; and (3) mathematical knowledge reasoning abilities are significantly degraded in current speech LLMs. 
\end{abstract}

\section{Introduction}

Recent advancements in large language models (LLMs) and multimodal large language models (MLLMs) have demonstrated impressive capabilities across various types of tasks, including question answering \cite{lewkowycz2022solving, guo2023images, shao2024deepseekmath}, code generation \cite{zheng2024makes, wei2024coinmath, liu2024exploring}, and speech understanding~\cite{gong2023joint, an2024funaudiollm, huang2025step}. Among this wide range of tasks, reasoning stands out as a core capability that underpins performance in complex downstream tasks. Consequently, considerable research has focused on evaluating and improving reasoning abilities across different modalities~\cite{shao2024deepseekmath, yang2024qwen2, xu2024llava, hurst2024gpt, thawakar2025llamav}.

While substantial progress has been made in evaluating reasoning in textual and visual domains, reasoning over spoken inputs remains comparatively underexplored. Initial studies~\cite{ghosh2024gama, sakshi2024mmau} have begun probing the reasoning capabilities of speech LLMs, but mainly in the context of basic comprehension or factual question answering, offering limited insight into multi-step reasoning processes such as Chain-of-Thought (CoT) reasoning. More recently, \citet{xie2025audio} introduced Audio-Reasoner, a large-scale audio-language model trained on an audio QA pair dataset, with a specific focus on CoT-style reasoning over spoken inputs. These developments mark important progress, but significant gaps remain, particularly in the domain of mathematical reasoning from speech.

Mathematical reasoning serves as a rigorous benchmark for evaluating model reasoning due to its reliance on logical step-by-step deduction, and sensitivity to fine-grained errors~\cite{Mishra2022Lila, lu2023mathvista, zhang2024mathverse}. Yet, most existing multimodal evaluations of mathematical reasoning focus heavily on visual-text settings~\cite{lu2023mathvista, zhang2024mathverse, wang2024measuring, qiao2024we}, with limited attention paid to spoken mathematical problem-solving. Some recent efforts have just begun exploring this direction. VoxEval~\cite{cui2025voxeval} evaluates spoken knowledge understanding, including mathematics, while StepEval-Audio-360~\cite{huang2025step} incorporates arithmetic questions as part of logical reasoning evaluations. However, a comprehensive benchmark for evaluating mathematical reasoning from spoken input is still missing. This oversight is critical, especially considering that spoken mathematical reasoning is highly relevant in real-world contexts such as classrooms, tutoring, and collaborative discussions. 

In this work, we address this gap by introducing Spoken Math Question Answering (Spoken-MQA), a novel benchmark designed to systematically evaluate the mathematical reasoning capabilities of speech-based models, including cascade models, i.e., Automatic Speech Recognition (ASR) followed by LLMs, and end-to-end speech LLMs, on multi-faceted spoken mathematical problems. Spoken-MQA includes pure arithmetic computations, single-step and multi-step contextual reasoning problems, as well as questions that require specialized domain knowledge in mathematics. All questions in Spoken-MQA are presented in unambiguous natural spoken language. To construct the benchmark, we incorporate both existing human-recorded spoken math problems and newly generated spoken data using a pipeline involving GPT-4o, human annotators, and text-to-speech (TTS) techniques. For the newly generated data, specifically, we first use GPT-4o to verbalize math problems that contain mathematical symbols and formulas. We then collaborate with human annotators and GPT-4o to filter for unambiguous verbalizations. The curated problems are finally synthesized into high-quality speech using a TTS system.

Through extensive experiments, we find that: (1) While some speech LLMs perform competitively on contextual reasoning tasks involving basic arithmetic, they still struggle with direct arithmetic questions, indicating a lack of genuine mathematical understanding despite surface-level competence. (2) Cascade models, which transcribe speech to text before reasoning, remain effective but suffer performance degradation due to ASR transcripts, especially for mathematical symbols and formula-heavy problems. Moreover, current LLMs in cascade models show a strong bias toward symbolic mathematical expressions, i.e., mathematical symbols and formulas in LaTeX forms, and have difficulty interpreting verbalized mathematical expressions, underscoring the need for improved natural spoken language understanding. (3) Most speech LLMs suffer from degraded mathematical reasoning capabilities, particularly on tasks requiring mathematical knowledge, as their training pipelines typically emphasize general conversational or audio understanding rather than mathematical domain-specific reasoning. These findings highlight key limitations in current speech-based models. We hope that Spoken-MQA serves as a valuable benchmark to drive further research in this emerging field.

Our contributions are summarized as follows:

\begin{itemize} 
\item We introduce the Spoken-MQA benchmark for spoken mathematical reasoning, enabling systematic evaluation of speech-based models across diverse math problem types and reasoning depths. 
\item We systematically compare current cascade models and end-to-end speech LLMs, revealing key limitations of current speech-based models on mathematical reasoning tasks, and provide detailed analysis and insights to guide future research on improving spoken mathematical reasoning.
\end{itemize}

\begin{figure*}[t]
    \centering
    \includegraphics[width=1.0\textwidth]{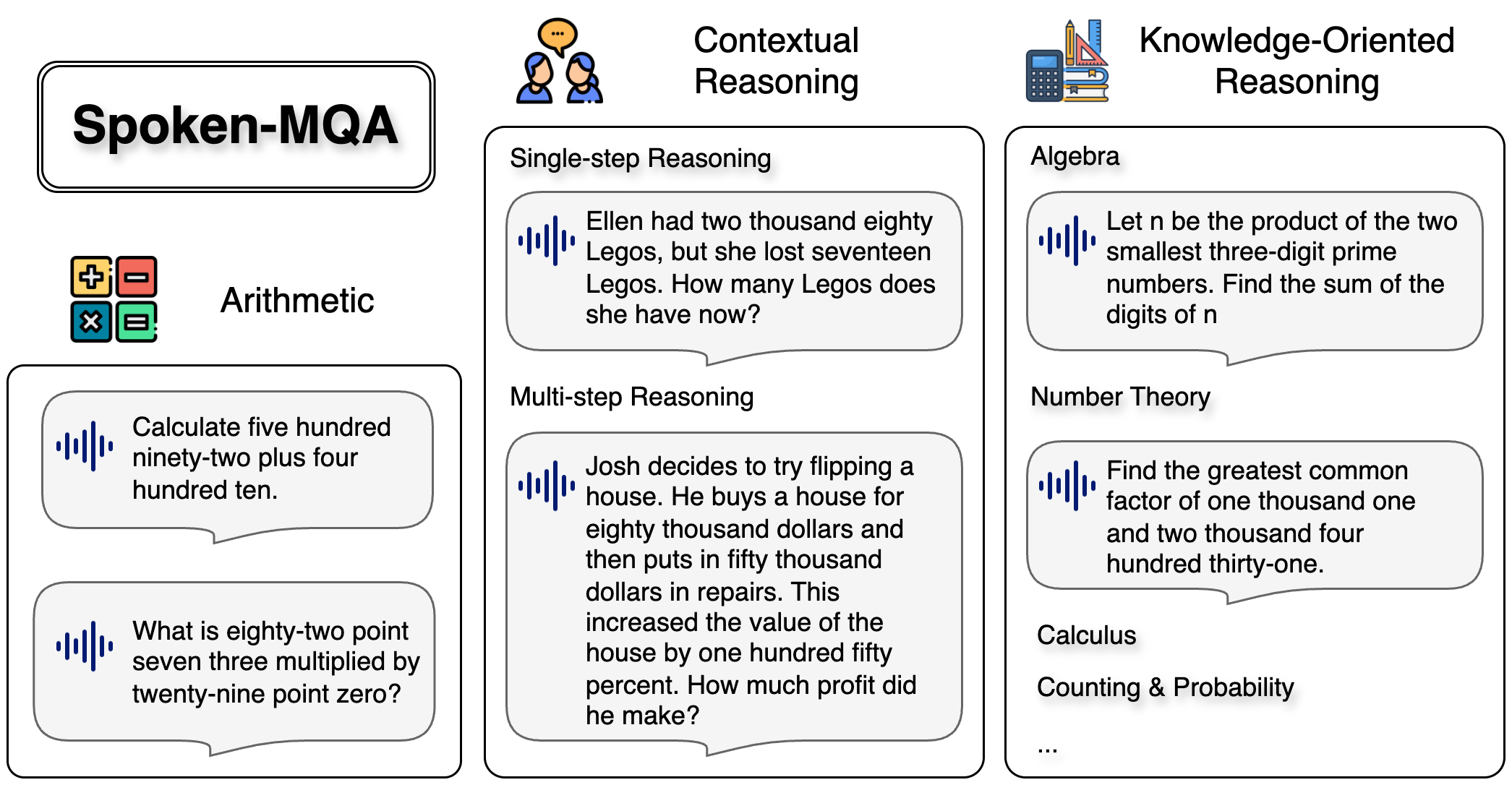}
    \caption{Overview of Spoken-MQA.}
    \label{fig:spoken-MQA}
\end{figure*}

\section{Related Work}

\textbf{Speech LLM}. Speech Large Language Models (Speech LLMs)~\cite{peng2024survey, cui2024recent} are emerging architectures that integrate speech understanding directly into the framework of large language models (LLMs). Different from cascade models, which first convert speech to text using Automatic Speech Recognition (ASR) before feeding it into LLMs, Speech LLMs process raw speech inputs end-to-end. They employ a speech encoder to extract representations from the speech signal, which are then mapped to the language model space via a projection module. This design reduces error propagation inherent in cascade pipelines and enables richer contextual comprehension by retaining paralinguistic cues. Recent models such as SpeechGPT~\cite{zhang2023speechgpt}, SALMONN~\cite{tang2023salmonn}, and Qwen-Audio~\cite{chu2023qwen, chu2024qwen2} exemplify this trend by leveraging large-scale pretraining and cross-modal alignment to jointly model speech, audio, and text. These models support a wide range of tasks, including speech recognition, translation, and spoken question answering.

\noindent \textbf{Reasoning Benchmark}. Models' reasoning abilities have been extensively evaluated in textual and visual modalities. In the textual domain, significant focus has been placed on mathematical reasoning, with benchmarks such as GSM8K \cite{cobbe2021training}, which targets grade-school level word problems, and MATH \cite{hendrycksmath2021}, which comprises competition problems at the high school level. Beyond mathematics, broader reasoning evaluations have emerged. For example, BIG-bench \cite{srivastava2022beyond}, MR-Bench \cite{zeng2024mr}, and ARB \cite{sawada2023arb} assess reasoning across diverse domains, including science, programming, logic, biology, chemistry, physics, and law. In the visual domain, benchmarks such as MathVista \cite{lu2023mathvista}, MathVerse \cite{zhang2024mathverse}, and MathVision evaluate mathematical reasoning involving images like diagrams and charts. Others, such as InfiMM-Eval \cite{han2023infimm}, MMMU \cite{yue2024mmmu}, EMMA \cite{hao2025can}, and VisuLogic \cite{xu2025visulogic}, further extend visual reasoning to complex real-world scenarios in subjects such as physics, chemistry, and general knowledge.

In contrast, speech and audio reasoning benchmarks are still in their early stages. MMAU \cite{sakshi2024mmau} and AIR-Bench \cite{yang2024air} assess general auditory understanding, including speech, environmental sounds, and music, through tasks like question answering and instruction following. StepEval-Audio-360 \cite{huang2025step} incorporates arithmetic within a logical reasoning framework. VoxEval \cite{cui2025voxeval} focuses on knowledge understanding from spoken inputs, including certain math-related queries. While these efforts represent important steps toward evaluating speech understanding and reasoning, they fall short in systematically assessing mathematical reasoning, an area well explored in text and vision modalities. This gap motivates our work: we propose a comprehensive benchmark designed to assess mathematical reasoning solely from spoken input.

\section{Spoken-MQA Benchmark Construction}

\subsection{Benchmark Taxonomy}

We introduce the \textbf{Spoken-MQA} benchmark, a curated collection of diverse textual math problems designed to evaluate spoken mathematical reasoning. As illustrated in Figure~\ref{fig:spoken-MQA}, the benchmark is structured into three categories based on the type of mathematical reasoning required:
(1) Arithmetic, which emphasizes direct numerical computation;
(2) Contextual Reasoning, involving single- or multi-step reasoning grounded in real-world scenarios; and
(3) Knowledge-Oriented Reasoning, which requires the application of mathematical domain-specific knowledge. 

\textbf{Arithmetic} consists of problems that focus on fundamental numerical operations, including addition, subtraction, multiplication, and division. A typical example in this category is: ``What’s 37.67 minus 75.7?'' The strong performance in arithmetic is a prerequisite for success in more complex mathematical reasoning tasks. The problems in the Arithmetic category are designed to require minimal contextual understanding and external knowledge in the mathematical domain, making them ideal for evaluating a model’s arithmetic ability. We source these problems from the SimpleMath subset of the Timers and Such dataset~\cite{lugosch2021timers}, utilizing recordings spoken by real human speakers. The selected problems cover both integer and decimal arithmetic, with certain questions requiring computations of moderate difficulty that involve numbers greater than four digits.

\textbf{Contextual Reasoning} contains everyday word problems that require interpreting short narratives and performing one or more arithmetic reasoning steps. These tasks evaluate a model’s ability to identify relevant quantities, reason over contextual information, and apply basic arithmetic operations. For example: ``Bobby ate 26 pieces of candy. Then, he ate 17 more. How many pieces of candy did Bobby eat?'' Problems in Contextual Reasoning category are sourced from datasets AddSub~\cite{hosseini2014learning, Mishra2022Lila}, SingleOp~\cite{roy2015solving, Mishra2022Lila}, SVAMP~\cite{patel2021nlp}, and GSM8K~\cite{cobbe2021training}, with increasing difficulty. AddSub focuses on single-step addition and subtraction problems. SingleOp extends to single-step multiplication and division. SVAMP introduces controlled variations to existing problems to test a model's sensitivity to subtle linguistic or structural changes, typically requiring up to two reasoning steps. GSM8K presents more challenging, multi-step grade-school problems that require advanced reasoning ability. To facilitate a more fine-grained evaluation, we divide this category into \textbf{single-step} and \textbf{multi-step} subsets based on the number of reasoning steps involved. Problems from AddSub and SingleOp are categorized as \textbf{single-step}, while those from GSM8K are assigned to the \textbf{multi-step} subset. For SVAMP, we apply a heuristic based on the number of arithmetic operations required in the solution to determine the reasoning complexity. This partition enables a more precise analysis of models’ performance to varying reasoning depths.

\textbf{Knowledge-Oriented Reasoning} comprises problems that require multi-step reasoning with specific mathematical knowledge, such as number theory, algebra, or modular arithmetic. For example: ``What is the smallest positive integer that leaves a remainder of 1 when divided by 2, 3, 4, 5, and 6, and is divisible by 7?''. Solving this math problem requires recognizing implicit constraints and applying mathematical knowledge like least common multiples and divisibility rules. We source these problems from the MATH dataset~\cite{hendrycksmath2021}, which includes high school competition-level problems across domains such as algebra, number theory, geometry, and probability. Compared to the previous two categories, problems in this category are more complex and contain extensive mathematical symbols and formulas written in LaTeX.

\subsection{Speech Conversion and Disambiguation}

To enable evaluation in the speech modality, we provide speech versions of math problems across all categories. While the problems in the Arithmetic category are originally collected in spoken form, other datasets are initially text-based. We synthesize their speech versions using the TTS technique.

\begin{figure}[ht]
    \centering
    \includegraphics[width=0.5\textwidth]{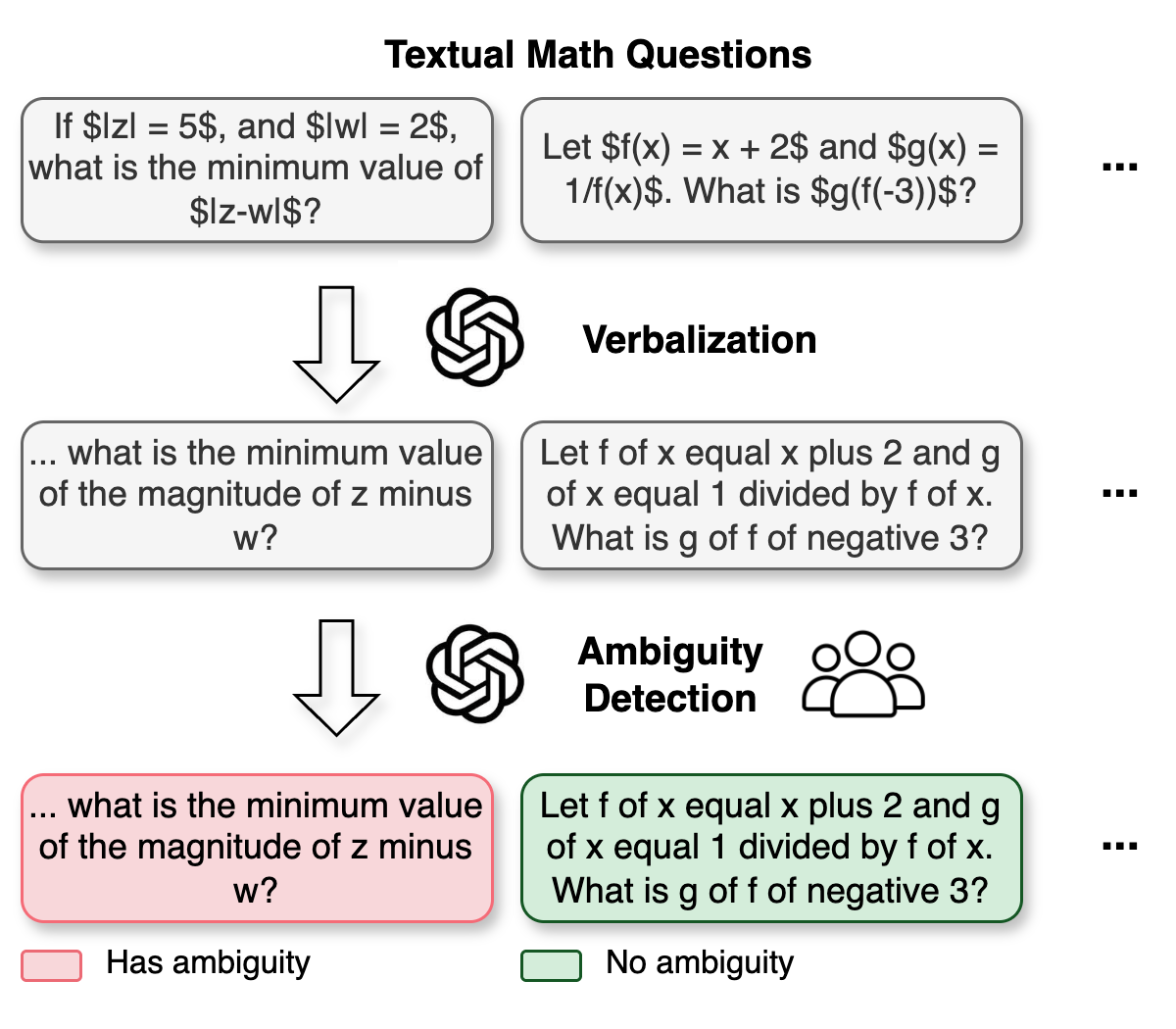}
        \caption{Pipeline for Generating and Filtering Unambiguous Verbal Math Questions. The process involves verbalizing math problems with LaTeX mathematical expressions using GPT-4o, followed by ambiguity detection through both GPT-4o and human verification.}
    \label{fig:data pipeline}
\end{figure}

However, converting complex math problems with heavy mathematical symbols and formulas in LaTeX into accurate and unambiguous speech introduces two major challenges. The first is accurate verbalization. For example, an expression like “9!” should be pronounced as “nine factorial,” not “nine exclamation mark,” which standard TTS systems may mispronounce. The second is verbal ambiguity, where a single spoken phrase can correspond to multiple written forms. For instance, “the magnitude of z minus w” could be interpreted as either $\vert z - w \vert$ or $\vert z \vert - w $. While such ambiguities are often resolved in textual modality, they remain ambiguous in speech alone. These issues are especially prevalent in the mathematical symbol- and formula-dense problems in the Knowledge-Oriented Reasoning category, whereas simpler problems in the Arithmetic and Contextual Reasoning categories are unaffected.

To better understand the extent of verbal ambiguity, we randomly sample 100 problems from the MATH dataset, which is the data source of the Knowledge-Oriented Reasoning category. We verbalize them using GPT-4o, where the instruction for GPT-4o is shown in Appendix~\ref{sec:appendix_prompt}. Then, human annotators assess their ambiguity based on the verbalized versions. Among these, 32 out of 100 problems are judged to be potentially ambiguous to humans without visual or text context. Representative examples and ambiguity types are detailed in Appendix~\ref{sec:appendix_ambiguity}. This analysis underscores the need for careful filtering and manual validation when constructing an unambiguous spoken math benchmark.

To address the two above-mentioned challenges, as illustrated in Figure~\ref{fig:data pipeline}, we first use GPT-4o to verbalize math problems and flag potentially ambiguous cases. We then filter out those identified as ambiguous and manually verify the remaining examples. Finally, a curated set of 500 unambiguous spoken math problems is created for the Knowledge-Oriented Reasoning category.

We use Coqui TTS\footnote{\url{https://github.com/coqui-ai/TTS}}, a high-performance deep learning-based TTS toolkit, to convert these curated problems into high-quality speech audio. The overall dataset statistics for Spoken-MQA are summarized in Table~\ref{tab:dataset}.

\begin{table}[h]
    \centering
    \large
    \begin{adjustbox}{width=1\linewidth, center}
    \begin{tabular}{c c l c c}
        \toprule
        \multicolumn{2}{c}{\textbf{Category}} & \textbf{Data Sources} & \textbf{\makecell{Num. \\ Questions}} & \textbf{\makecell{Avg. Length \\ (Words)}} \\
        \midrule
        \multicolumn{2}{c}{Arithmetic} & SimpleMath & 273 & 4.5 \\
        \midrule
        \multirow{2}{*}[-3ex]{\centering\makecell[l]{Contextual\\Reasoning}} 
            & Single-step & \makecell[l]{AddSub\\SingleOp\\SVAMP} & 594 & 29.3 \\
        \cmidrule{2-5}
            & Multi-step & \makecell[l]{SVAMP\\GSM8K} & 1402 & 51.5 \\
        \midrule
        \multicolumn{2}{c}{Knowledge-Oriented Reasoning} & MATH & 500 & 36.1 \\
        \bottomrule
    \end{tabular}
    \end{adjustbox}
    \caption{Statistics of Spoken-MQA Benchmark} \label{tab:dataset}
\end{table}

\section{Experiments}
\begin{table*}[h]
    \centering
    \begin{adjustbox}{width=0.9\textwidth,center}
    \begin{tabular}{lccccc}
        \toprule
        \multirow{2}{*}{Model} & \multirow{2}{*}{Arithmetic} & \multicolumn{2}{c}{\makecell{ Contextual \\ Reasoning}} & \multirow{2}{*}[1.3ex]{\makecell{Knowledge\\-oriented\\Reasoning}} & \multirow{2}{*}{Avg.}\\
        &  & Single-step & Multi-step &  & \\

        \midrule
        \multicolumn{6}{c}{Cascade Model} \\
        \midrule
        Whisper-Mistral-7B-Instruct & 19.8 & 76.9 & 49.6 & 11.8 & 39.5 \\
        Whisper-Llama-3.1-8B-Instruct & 67.0 & 87.2 & 77.6 & 36.4 & 67.0 \\ 
        Whisper-Gemma-2-9B-Instruct & 66.7 & 88.2 & 81.7 & 46.6 & 70.8 \\
        Whisper-Qwen2.5-7B-Instruct & 70.0 & 81.0 & 68.9 & 57.8 & 69.4 \\
        \midrule
        Whisper-Mathstral-7B & 68.5 & 87.4 & 76.3 & 45.2 & 69.3 \\
        Whisper-Deepseek-Math-7B-instruct & 61.2 & 85.2 & 78.2 & 44.8 & 67.4 \\
        Whisper-Qwen2.5-Math-7B-Instruct & \textbf{77.3} & \textbf{88.0} & \textbf{86.2} & \textbf{72.4} & \textbf{81.0} \\
        \midrule
        \multicolumn{6}{c}{End-to-End Speech LLM} \\
        \midrule
        GPT-4o-audio & \textbf{71.8} & \textbf{90.7} & \textbf{85.9} & \textbf{67.6} & \textbf{79.0} \\
        \midrule
        Qwen2-Audio-7B-Instruct & 46.9 & 56.2 & 19.2 & 5.8 & 32.0 \\
        Audio-Reasoner-7B & 36.3 & 58.6 & 28.6 & 6.6 & 32.5 \\
        MERaLiON-AudioLLM-9B & 50.5 & 78.5 & 39.3 & 8.8 & 44.3 \\
        Ultravox-llama-3.1-8B & 45.4 & 87.7 & 78.5 & 36.6 & 62.0 \\
        Phi-4-multimodal-6B-instruct & 38.8 & 86.4 & 77.7 & 27.2 & 57.5 \\
        FT-Phi-4-multimodal-6B-instruct & 56.4 & 87.5 & 80.1 & 52.2 & 69.0 \\
        \bottomrule
    \end{tabular}
    \end{adjustbox}
    \caption{Accuracy (\%) of models on the Spoken-MQA benchmark. The best-performing cascade model and end-to-end speech LLM are highlighted in bold.}
    \label{tab:model_performance}
\end{table*}

\subsection{Experimental Setup}

We examine two types of Speech models: cascade models and end-to-end speech LLMs.

\textbf{Cascade Model}. In the cascade approach, speech inputs are first transcribed into text using Whisper-Large-V3 \cite{radford2023robust}, a state-of-the-art ASR model known for its robustness and accuracy. The transcriptions are then passed to text-based LLMs for evaluation. We consider the following generic LLMs:  \texttt{Mistral-7B-Instruct} \cite{jiang2023identifying}, \texttt{LLaMA-3.1-8B-Instruct} \cite{grattafiori2024llama}, \texttt{Gemma-2-9B-Instruct} \cite{team2024gemma}, and \texttt{Qwen2.5-7B-Instruct} \cite{qwen2.5}. In addition, we include math-specialized LLMs: \texttt{Mathstral-7B} \cite{mathstral2024}, \texttt{Qwen2.5-Math-7B-Instruct} \cite{yang2024qwen25mathtechnicalreportmathematical}, \texttt{DeepSeek-Math-7B-Instruct} \cite{shao2024deepseekmath}. For clarity, we denote each cascade model as \texttt{Whisper-[LLM Name]}.

\textbf{Speech LLM}. These models process speech inputs directly, integrating both speech and language understanding within a unified framework. We evaluate the following state-of-the-art speech LLMs:

\begin{itemize}
    \item \texttt{Qwen2-Audio-7B-Instruct}~\cite{chu2024qwen2} integrates the Whisper-large-v3 model as the audio encoder with Qwen2-7B language model to support tasks such as voice chat and audio analysis.

    \item \texttt{Audio-Reasoner-7B} \cite{xie2025audio} is designed to enhance reasoning capabilities over audio inputs. It is built upon Qwen2-Audio-Instruct and focuses on improving the model's ability to perform complex audio reasoning tasks.

    \item \texttt{MERaLiON-AudioLLM-9B}~\cite{he2024meralion} combines a tuned Whisper-large-v2 with the SEA-LION V3 language model (Llama-3.1-8B-Instruct), connected through an adaptor module that aligns the speech or audio embeddings with the text embedding space.

    \item \texttt{Ultravox-llama-3.1-8B}\footnote{\url{https://github.com/fixie-ai/ultravox}} is built upon a pretrained Llama3.1-8B-Instruct language model and a Whisper-large-v3-turbo audio encoder.

    \item \texttt{Phi-4-Multimodal-6B-Instruct} \cite{abouelenin2025phi} integrates the pretrained Phi-4-Mini-Instruct language model with advanced encoders and adapters for vision and speech, allowing multiple inference modes combining various modalities, including speech.

    \item \texttt{GPT-4o-Audio} \cite{hurst2024gpt} is a proprietary model. It serves as a benchmark for evaluating the performance gap between open-source and proprietary speech LLMs.
\end{itemize}

In addition, we fine-tune \texttt{Phi-4-Multimodal\allowbreak-6B-Instruct}, referred to as \texttt{FT-Phi-4\allowbreak-Multimodal-Instruct}, using 500k samples from the OpenMathInstruct-1 dataset~\cite{toshniwal2024openmath}. The training details are listed in Appendix~\ref{sec:appendix_training}. To reduce ambiguity in spoken input and minimize annotation effort when forming the training set, we filter out problems containing complex mathematical expressions. 500k selected textual questions are subsequently synthesized into speech via Coqui TTS.

All models are evaluated in a zero-shot CoT setting, where prompts are appended with "Please reason step by step." The evaluation metric is accuracy, determined by comparing the model's final answer to the ground truth.
 
\subsection{Experimental Result \& Analysis}

Table~\ref{tab:model_performance} presents the evaluation results of the aforementioned models on the Spoken-MQA benchmark. Among cascade models, \texttt{Whisper-Qwen2.5-7B-Instruct} achieves the best overall performance due to the strong mathematical reasoning capabilities of its underlying text-based LLM. For end-to-end speech LLMs, the proprietary \texttt{GPT-4o-audio} model achieves the best results. Overall, state-of-the-art open-source speech LLMs underperform compared to cascade models. Among them, only \texttt{Phi-4-multimodal-instruct} and \texttt{Ultravox} demonstrate relatively strong performance in the Contextual Reasoning category. Notably, both \texttt{Ultravox} and \texttt{Whisper-Llama-3.1-8B-Instruct} are based on the Llama-3.1-8B-Instruct language model, and they achieve similar performance in Contextual and Knowledge-Oriented Reasoning. However, in the Arithmetic category, most speech LLMs lag significantly behind the cascade models.

In the following subsections, we conduct a detailed analysis of model performance from several perspectives: the arithmetic gap, the impact of input format on LLMs, and the effects of domain-specific fine-tuning.

\subsubsection{Arithmetic Gap}
 
A clear performance gap exists between cascade models and open-source speech LLMs in the Arithmetic category. Intriguingly, although both the Arithmetic and Contextual Reasoning categories involve arithmetic operations on numbers ranging from one to three digits, speech LLMs such as \texttt{Phi-4-Multimodal-Instruct} and \texttt{Ultravox}, which perform competitively on the Contextual Reasoning category, exhibit poor performance on the Arithmetic category.

To investigate the factors contributing to this discrepancy, we further analyze model performance with respect to arithmetic complexity. Specifically, we partition the samples in the Arithmetic category into two subsets: short-digit problems, which involve numbers with up to three digits (comparable to those in the Contextual Reasoning category), and long-digit problems, which involve numbers with four or more digits.

\begin{figure}[h]
    \centering
    \includegraphics[width=0.5\textwidth]{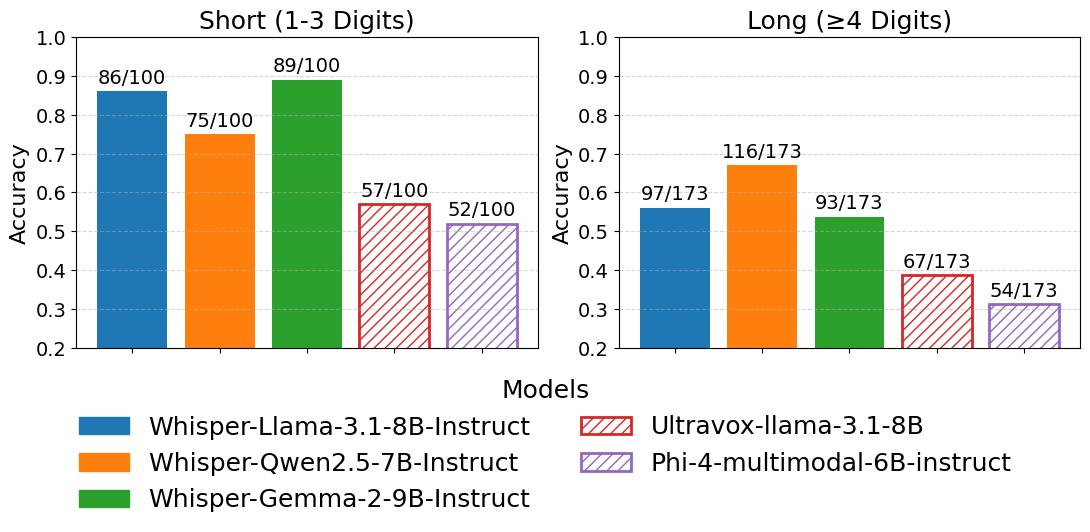}
        \caption{Model Accuracy on Short vs. Long Digit Length in Arithmetic}
    \label{fig:arith analysis}
\end{figure}

The results, presented in Figure~\ref{fig:arith analysis}, include selected cascade models and two speech LLMs, \texttt{Ultravox} and \texttt{Phi-4-multimodal-instruct}, for their relatively strong performance in Contextual Reasoning. Speech LLMs underperform both on short-digit and long-digit arithmetic compared to cascade models. This suggests that their overall poor performance in the Arithmetic category cannot be attributed solely to the challenge of handling large numbers. Instead, it may also stem from limitations in the speech encoder or the arithmetic capabilities of these models. Furthermore, as speech LLMs such as \texttt{Ultravox} and \texttt{Phi-4-Multimodal-Instruct} underperform on short-digit arithmetic, their competitive performance on Contextual Reasoning tasks may be driven by superficial pattern matching, that is, relying on surface-level cues rather than demonstrating a true understanding of underlying mathematical computation. Investigating whether these models genuinely comprehend the logic behind arithmetic operations or merely exploit shallow correlations remains an open direction for future research.

\subsubsection{Impact of Input Format on LLMs}

\begin{table*}[h]
    \centering
    \begin{adjustbox}{width=0.99\textwidth,center}
    \begin{tabular}{ccllllllll}
        \toprule

                \multirow{2}{*}{Model} & \multirow{2}{*}{Input} & \multirow{2}{*}{Arithmetic} & \multicolumn{2}{c}{\makecell{ Contextual \\ Reasoning}} & \multirow{2}{*}[1.3ex]{\makecell{Knowledge\\-oriented\\Reasoning}} & \multirow{2}{*}{Avg.}\\
        &  &  & Single-step & Multi-step &  & \\
        \midrule
        \multirow{3}{*}{Llama-3.1-8B-Instruct} 
        & ASR Transcript & 67.0 & 87.2 & 77.6 & 36.4 & 67.1 \\
        & Verbalized Text & - & - & -  & 43.4 \textcolor{green}{(+7.0)} & - \\
        & Ground-Truth Text & 70.7 \textcolor{green}{(+3.7)} & 88.6 \textcolor{green}{(+1.4)} & 81.6 \textcolor{green}{(+4.0)} & 53.2 \textcolor{green}{(+16.8)} & 73.5 \textcolor{green}{(+6.4)} \\
        \midrule
        \multirow{3}{*}{Gemma-2-9b-Instruct} 
        & ASR Transcript & 66.7 & 88.2 & 81.7 & 46.6 & 70.8 \\
        & Verbalized Text & - & - & - & 52.2 \textcolor{green}{(+5.6)} & - \\
        & Ground-Truth Text & 71.1 \textcolor{green}{(+4.4)} & 87.9 \textcolor{red}{(-0.3)} & 84.8 \textcolor{green}{(+3.1)} & 57.0 \textcolor{green}{(+10.4)} & 75.2 \textcolor{green}{(+4.4)} \\
        \midrule
        \multirow{3}{*}{Qwen2.5-7B-Instruct} 
        & ASR Transcript & 70.0 & 81.0 & 68.9 & 57.8 & 69.4 \\
        & Verbalized Text & - & - & -  & 69.8 \textcolor{green}{(+12.0)} & - \\
        & Ground-Truth Text & 72.2 \textcolor{green}{(+2.2)} & 82.3 \textcolor{green}{(+1.3)} & 73.0 \textcolor{green}{(+4.1)} & 75.4 \textcolor{green}{(+17.6)} & 75.7 \textcolor{green}{(+6.3)} \\
        \midrule
        \multirow{3}{*}{Deepseek-Math-7B-Instruct} 
        & ASR Transcript & 61.2 & 85.2 & 78.2 & 44.8 & 67.3 \\
        & Verbalized Text & - & - & -  & 52.0 \textcolor{green}{(+7.2)} & - \\
        & Ground-Truth Text & 62.6 \textcolor{green}{(+1.4)} & 86.9 \textcolor{green}{(+1.7)} & 81.2 \textcolor{green}{(+3.0)} & 53.6 \textcolor{green}{(+8.8)} & 71.1 \textcolor{green}{(+3.8)} \\
        \midrule
        \multirow{3}{*}{Qwen2.5-Math-7B-Instruct} 
        & ASR Transcript & 77.3 & 88.0 & 86.2 & 72.4 & 81.0 \\
        & Verbalized Text & - & - & - & 83.8 \textcolor{green}{(+11.4)} & - \\
        & Ground-Truth Text & 78.4 \textcolor{green}{(+1.1)} & 90.9 \textcolor{green}{(+2.9)} & 91.2 \textcolor{green}{(+5.0)} & 89.0 \textcolor{green}{(+16.6)} & 87.4 \textcolor{green}{(+6.4)} \\
        \bottomrule
    \end{tabular}
    \end{adjustbox}
    \caption{Accuracy across three input formats (Whisper transcript, Verbalized Text, and Ground Truth Text) to LLMs. Performance change from ASR Transcript is shown inline with \textcolor{green}{↑} for improvement and \textcolor{red}{↓} for degradation.}
    \label{tab:asr}
\end{table*}

Cascade models, which transcribe speech into text via ASR systems before passing it to LLMs, demonstrate strong performance on the Spoken-MQA benchmark. This highlights that cascade-based approaches currently offer an effective solution for speech-based instruction tasks in mathematical reasoning. To better understand how input formats influence LLM performance, we compare results using ASR-transcribed input against clean, ground-truth text. Additionally, for problems in the Knowledge-Oriented Reasoning category, which often contain extensive mathematical symbols and formulas written in LaTeX, we assess performance using verbalized text, where mathematical symbols and formulas are written in natural spoken language.

Model performances are summarized in Table~\ref{tab:asr}. The results reveal several key trends. First, all models generally achieve higher accuracy when provided with clean, ground-truth text inputs compared to ASR-transcribed ones. This trend is especially observable in the Knowledge-oriented Reasoning category, which frequently includes complex mathematical expressions. In contrast, tasks like Arithmetic and Single-step Contextual Reasoning, which contain simpler language and fewer mathematical expressions, are less affected by transcription errors. Nevertheless, the observed small performance gap indicates that despite recent advances in ASR technology, transcription errors can still propagate and negatively affect the downstream performance of cascade models in mathematical reasoning tasks.

Second, while verbalized text preserves the semantic content of the original textual questions, it consistently yields lower accuracy compared to ground-truth text inputs. This performance gap reveals a significant limitation of current LLMs in the context of speech-based mathematical reasoning: LLMs exhibit a strong bias toward symbolic mathematical expressions (e.g., “$x^2 + y^2 = 1$”), and struggle when mathematical symbols and formulas are presented in natural spoken language (e.g., “x squared plus y squared equals one”). This highlights the need for further research to reduce the symbolic bias of LLMs and enhance the model for more robust natural spoken language understanding.

\subsubsection{Domain-specific Fine-tuning}

Notably, as shown in Table~\ref{tab:model_performance}, \texttt{Ultravox}, which is built on Llama-3.1-8B-Instruct, performs comparably to its cascade counterpart, \texttt{Whisper-Llama-3.1-8B-Instruct}, on most Spoken-MQA tasks except the Arithmetic category. This highlights the need for proper pretraining and cross-modal alignment strategies in speech LLMs to retain the reasoning capabilities inherited from their text-based LLMs. However, the specific training methodology behind \texttt{Ultravox} remains undisclosed.

Other speech LLMs significantly underperform compared to cascade models. This notable performance gap highlights a major limitation of current speech LLMs: they are not explicitly optimized for domain-specific reasoning, or such capabilities may be degraded during training, as their training typically focuses on general conversational, audio, and speech understanding tasks.

Supporting this, \texttt{FT-Phi-4-multimodal-instr\allowbreak-uct}, fine-tuned on a 500k-sample speech instruction dataset in the mathematical domain, shows substantial performance gains in the Arithmetic and Knowledge-oriented Reasoning categories, with more modest improvements in Contextual Reasoning. These results suggest that fine-tuning with domain-specific spoken data can significantly enhance a model’s ability to comprehend and solve spoken math problems, especially those involving arithmetic operations and knowledge-required reasoning. The relatively smaller gain in Contextual Reasoning may be due to the already strong baseline performance of the model before fine-tuning, comparable to most cascade models in Table~\ref{tab:model_performance} and even approaching the performance of \texttt{GPT-4o-audio}.


\section{Conclusion}

We introduce Spoken-MQA, a benchmark for evaluating the mathematical reasoning abilities of speech-based models, including cascade models (ASR + LLMs) and end-to-end speech LLMs. Covering diverse problem types from arithmetic to contextual and knowledge-oriented reasoning, Spoken-MQA fills a critical gap in assessing spoken mathematical understanding. Experimental results show that current end-to-end speech LLMs generally lag behind cascade models, especially in arithmetic and knowledge-intensive tasks. Additionally, LLMs in cascade systems exhibit a strong preference for symbolic mathematical LaTeX expressions over ASR transcripts or verbalized math. Lastly, domain-specific fine-tuning proves effective, notably boosting performance on arithmetic and knowledge reasoning tasks. These findings highlight the need to improve models' understanding of verbal mathematical expressions and to incorporate domain-adaptive training, paving the way for more capable speech-based reasoning models.

\section*{Limitations}

The Contextual Reasoning and Knowledge-Oriented Reasoning subsets in Spoken-MQA are generated using a TTS system. While this approach ensures consistency and scalability, the synthesized speech may introduce biases during model evaluation, as it lacks the natural variability and imperfections found in human speech. Such biases could affect the generalizability of model performance in real-world applications. Future work is needed for more human-spoken data across all subsets to better reflect authentic speech patterns and to ensure a more realistic and comprehensive evaluation of speech-based mathematical reasoning.

\section*{Acknowledgments}

\bibliography{custom}

\appendix

\begin{figure*}[h]
    \centering
    \includegraphics[width=0.8\textwidth]{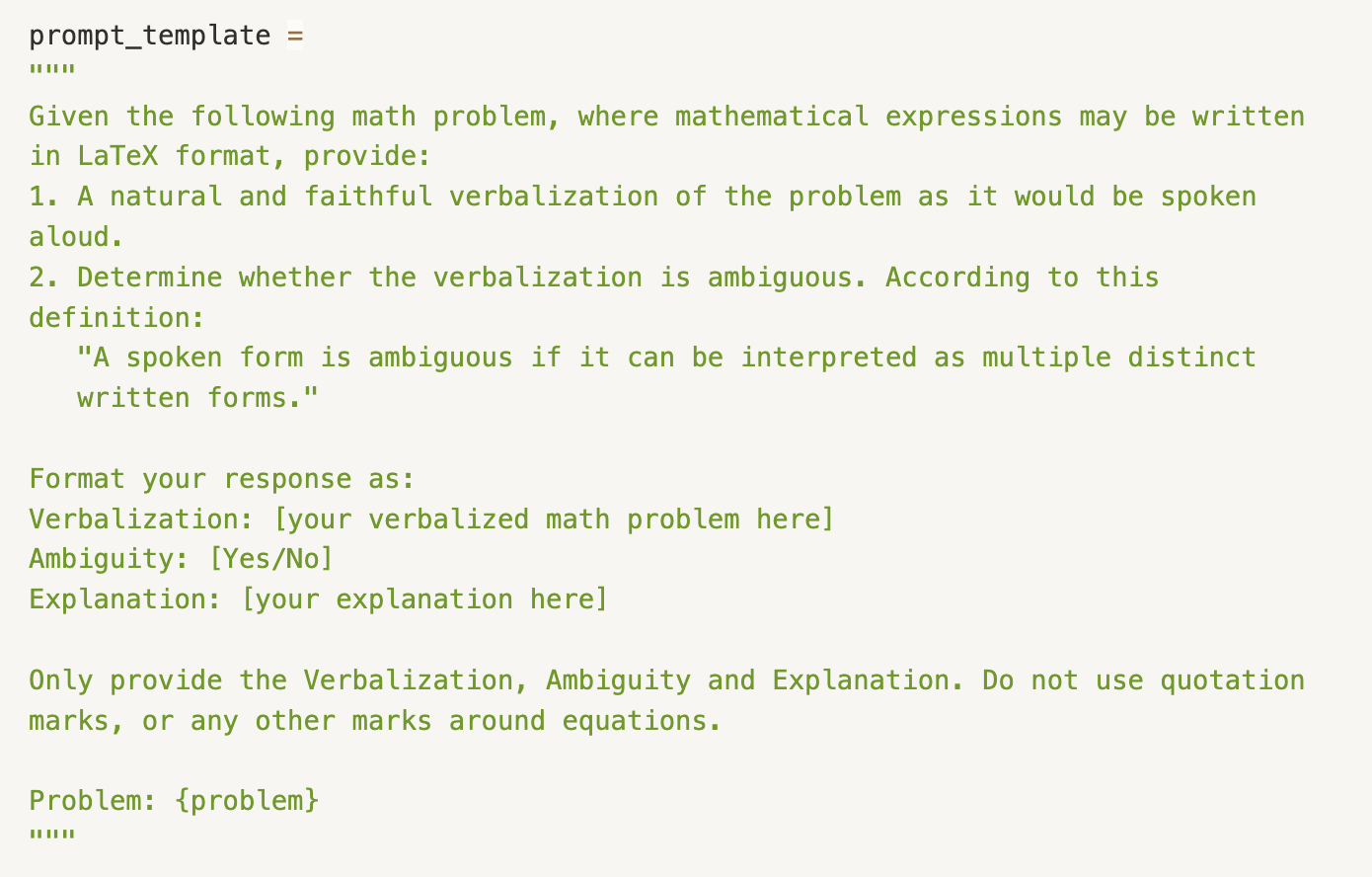}
    \caption{Instruction for generating verbalized math questions and assessing their ambiguity}
    \label{fig:appendix_instruction}
\end{figure*}

\section{Prompts for Generating Verbalized Math Questions}
\label{sec:appendix_prompt}
Figure~\ref{fig:appendix_instruction} presents the instructions used to generate verbalized math questions and assess their ambiguity.

\section{Ambiguous Examples}
\label{sec:appendix_ambiguity}

Figure~\ref{tab:ambiguous_verbal} presents examples of problems that exhibit ambiguity after being verbalized.

\begin{table*}[ht]
\centering
\renewcommand{\arraystretch}{1.4}
\begin{tabular}{|p{5cm}|p{5cm}|p{5cm}|}
\hline
\textbf{Original Math Problem} & \textbf{Verbalized Version} & \textbf{Explanation of Ambiguity} \\
\hline
Find the maximum value of $ f(x)=\sqrt{8x - x^2} - \sqrt{14x-x^2 - 48} $ & 
Find the maximum value of f of x equals the square root of eight x minus x squared, minus the square root of fourteen x minus x squared minus forty-eight. & Unclear where the square root ends. For example, could mean either 
\(\sqrt{14x - x^2 - 48}\) or \(\sqrt{14x} - x^2 - 48\). \\
\hline
Simplify: \(92 - 45 \div (3 \times 5) - 5^2\) & Simplify ninety-two minus forty-five divided by the quantity three times five minus five squared &  
Ambiguous whether "minus five squared" is part of the denominator or a separate term; could be interpreted as \(\frac{45}{3 \times 5 - 5^2}\) or \(\frac{45}{3 \times 5} - 5^2\). \\
\hline
Find the largest value of $c$ such that $\frac{c^2 + 6c -27}{c-3} +2c = 23$. & Find the largest value of c such that the fraction c squared plus 6c minus 27 over c minus 3, plus 2c, equals 23. & Ambiguity in whether “plus 2c” is part of the numerator/denominator or a separate term; could be interpreted as \(\frac{c^2 + 6c - 27}{c - 3 + 2c}\) or \(\frac{c^2 + 6c - 27}{c - 3} + 2c\). \\
\hline
Let $f(x)$ be the polynomial $f(x)=x^7-3x^3+2.$ If $g(x) = f(x + 1)$, what is the sum of the coefficients of $g(x)$? & Let f of x be the polynomial f of x equals x to the power of seven minus three x to the power of three plus two. If g of x equals f of x plus one, what is the sum of the coefficients of g of x? & Ambiguous whether “plus one” refers to \( f(x + 1) \) or \( f(x) + 1 \). \\
\hline

\end{tabular}
\caption{Examples of ambiguous verbalized math problems}
\label{tab:ambiguous_verbal}
\end{table*}

\section{Training Strategy}
\label{sec:appendix_training}

The Phi-4 Multimodal Instruct model is trained by a parameter-efficient fine-tuning strategy using Low-Rank Adaptation (LoRA). The base model and speech encoder parameters remain frozen, and only the audio project and LoRA\_A adapters are trained. We set the number of training epochs to 3 and use a learning rate of 4e-5. 

\end{document}